\documentclass{article}

\usepackage{iclr2027_conference,times}

\usepackage{amsmath,amsfonts,bm}

\def\eqref#1{equation~\ref{#1}}

\def\1{\bm{1}}

\DeclareMathAlphabet{\mathsfit}{\encodingdefault}{\sfdefault}{m}{sl}
\SetMathAlphabet{\mathsfit}{bold}{\encodingdefault}{\sfdefault}{bx}{n}

\usepackage{hyperref}
\usepackage{url}
\usepackage{graphicx}
\usepackage{booktabs}
\usepackage{tabularx}
\usepackage{array}
\usepackage{xcolor}
\usepackage{microtype}
\usepackage{mathtools}
\usepackage{amsthm}
\usepackage{enumitem}
\usepackage{float}
\usepackage{placeins}
\usepackage{amssymb}
\hypersetup{
hypertexnames=false,
pdftitle={Beyond Multimodal Alignment: Shared Physical Representations Across Sensors and Action Orders},
pdfauthor={Kaizhen Tan, Xin Xu, Siru Tao, Yixiao Li, Hanzhe Hong, Yang Feng, Heqing Du}
}

\newcommand{\Qfit}{\mathcal{Q}_{\mathrm{fit}}}
\newcommand{\Qdev}{\mathcal{Q}_{\mathrm{dev}}}
\newcommand{\Qtest}{\mathcal{Q}_{\mathrm{test}}}
\newcommand{\Qbridge}{\mathcal{Q}_{\mathrm{bridge}}}
\newcommand{\Rop}{\mathcal{R}}
\newcommand{\ES}{\operatorname{ES}}

\newtheorem{proposition}{Proposition}

\title{Beyond Multimodal Alignment:\\
Shared Physical Representations\\
Across Sensors and Action Orders}

\author{\textbf{Kaizhen Tan\textsuperscript{1,2}, Xin Xu\textsuperscript{2}, Siru Tao\textsuperscript{2}, Yixiao Li\textsuperscript{2}}\\
\textbf{Hanzhe Hong\textsuperscript{2}, Yang Feng\textsuperscript{3}, Heqing Du\textsuperscript{3}}\\
\normalfont\textsuperscript{1}New York University, New York, NY, USA\\
\normalfont\textsuperscript{2}Carnegie Mellon University, Pittsburgh, PA, USA\\
\normalfont\textsuperscript{3}Columbia University, New York, NY, USA}

\iclrfinalcopy
\begin{document}

\maketitle

\begin{abstract}
Multimodal world models are often evaluated by whether different sensors produce similar representations. However, similar representations do not necessarily imply that the models make the same physical predictions, or that those representations can be reused when actions are combined in a new order. We study both questions through the physical responses predicted by a model. We first use the Cluster Haptic dataset to ask whether audio and acceleration can independently recover the behavior of the same surface from different observations. Predictions from the two sensors are substantially closer for the same surface than for different surfaces, with a $4.5\times$ gap on average, while both also outperform an average-surface prediction. We then show that this agreement alone does not determine how familiar actions should compose. In a controlled elastoplastic system, shared step dynamics fit observed programs less accurately than a whole-program predictor but generalize better to unseen action orders, with the ranking reversing on both held-out transitions across three independent initializations. Fusing free-decay and hysteresis observations further improves prediction, with diagonal Gaussian beliefs yielding the lowest errors. Together, these results distinguish cross-sensor consistency, multimodal fusion, and generalization to new action orders as separate questions in evaluating multimodal physical representations.
\end{abstract}

\section{Introduction}

Multimodal world models use shared representations to connect sensory observations with predictions of physical behavior~\citep{shang2026phizero,tan2026latent,jin2026tacwam}. Different sensors can observe the same entity through very different signals. A surface scanned under a controlled contact, for example, produces both audio and acceleration, each varying with scan speed, direction, and force~\citep{eguchi2026cluster}. If these observations support a shared physical representation, their predictions should preserve the surface's behavior under additional scans. This motivates evaluating a representation through the physical responses it predicts when its sensory source changes.

Embedding alignment measures proximity between representations, while reconstruction and attribute probes assess decoding and recoverable physical distinctions. Response prediction adds a behavioral test: do different sensory observations lead to consistent predictions for the same entity? Predictive-state and function-representation methods motivate describing a system through its responses~\citep{littman2001predictive,ingebrand2024zeroshot}. Predictions that agree only near a population average carry little information about a particular surface, and agreement between two predictions does not establish their accuracy. We therefore compare same-entity responses with wrong-entity and population controls, and separately evaluate error against measured responses.

\begin{figure}[t]
    \centering
    \includegraphics[width=\linewidth]{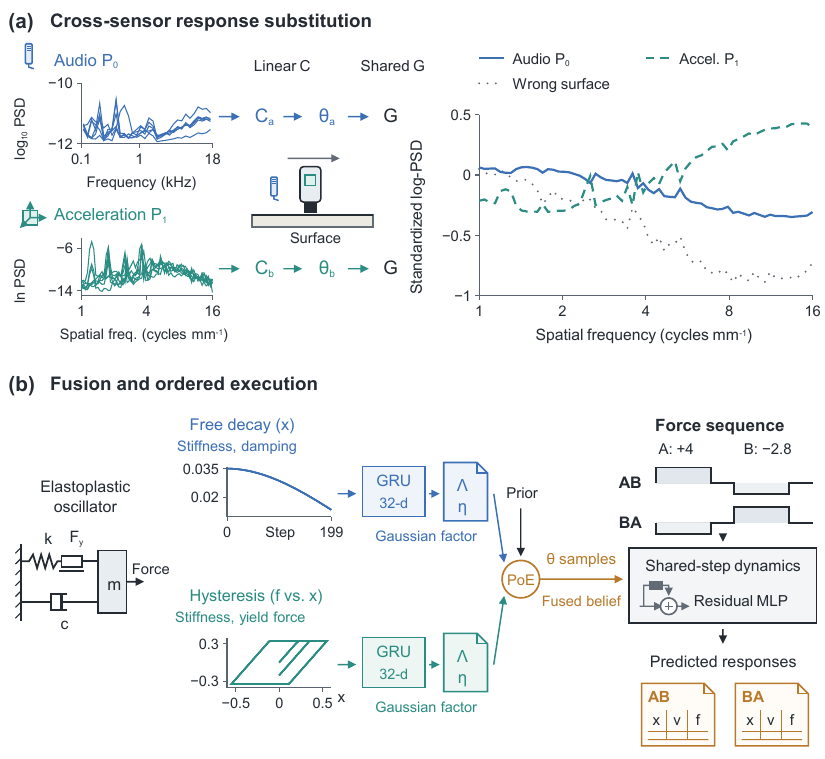}
    \caption{\textbf{Shared response coordinates across sensors and action orders.}
(a) Independently fitted audio and acceleration compilers infer separate coordinates,
compared through a shared frozen response model. The curves illustrate one query's
normal acceleration spectrum.
Four branches cover both cross-panel orientations.
(b) Free decay is invariant to yield force $F_y$, and settled hysteresis to damping $c$.
GRUs infer Gaussian factors $(\Lambda,\eta)$, combined with the prior by a product of
experts (PoE). Fused coordinate samples and force inputs enter shared step dynamics.
Response sheets depict the post-action output format $(x,v,f)$; physical drawings
and parameter sheets are schematic.}
    \label{fig:overview}
\end{figure}

We instantiate this approach on Cluster Haptic using audio and acceleration recorded under non-overlapping sets of scan conditions. A sensor-specific mapping, which we call a compiler, infers a coordinate in a common response chart. A shared response model predicts additional scans (Figure~\ref{fig:overview}a). The chart and response model are learned from training surfaces and frozen before the four modality--panel compilers are fitted independently, without entity identifiers or a cross-modal matching loss. This common response supervision makes the predictions comparable while allowing each sensor to infer a new surface's behavior from its own observations.

Using a physical representation for sequential prediction requires a model of how each input changes the state encountered by the next input. We examine this requirement in an elastoplastic system, where familiar force pulses occur in new orders (Figure~\ref{fig:overview}b). An ambiguity construction separates recovery of entity information from identification of an unseen composition law. We compare shared step dynamics with whole-program prediction and evaluate fusion of complementary observations. Each study uses its own chart and executor to test a different aspect of shared physical representation.

We evaluate whether shared physical representations preserve entity-specific responses across sensors and support new action orders. The real-data study changes the sensory source while holding the response model fixed. The controlled study compares executors with fixed entity coordinates, then evaluates complementary observations through a frozen executor. On Cluster Haptic, all four branches predict measured responses better than a population baseline, and every test surface has a smaller same-surface distance than its mean wrong-surface distance. In the controlled system, the whole-program predictor fits observed programs better, but shared step dynamics generalize better to both unseen transitions in all three initializations. Complementary fusion further improves prediction, with the diagonal belief performing best among the tested formulations.

\section{Related Work}

Physical world models learn representations for transition prediction, mechanical-property recovery, and tactile actions~\citep{shang2026phizero,tan2026latent,jin2026tacwam}. Cross-modal interchangeability asks whether different sensors support a common task. Connect--Collapse--Corrupt learns cross-modal tasks from unimodal data, and function-contrastive learning connects views of a common function~\citep{zhang2024connect,gondal2021function}. Partial-view causal representation learning examines shared content under partial observability~\citep{yao2024multiview}. We fit sensor mappings separately to a fixed response model and compare predictions beyond their input observations. Wrong-entity controls and measured responses test specificity and accuracy.

Describing a physical entity by its responses follows a predictive view of representation. Predictive-state models and causal states characterize systems through possible future observations~\citep{littman2001predictive,downey2017psrnn,shalizi2001computational}. Function Encoders, neural processes, and DeepONet learn function representations from input--output observations~\citep{ingebrand2024zeroshot,garnelo2018neural,lu2021deeponet}. We build on this view by comparing the physical responses predicted from different sensory observations.

Gaussian products of experts fuse evidence~\citep{hinton2002products,wu2018multimodal,sutter2021generalized}, while hidden-parameter state-space models represent uncertainty over task properties~\citep{shaj2022hidden}. Compositional-generalization studies evaluate new combinations of familiar components~\citep{lake2017generalization,hupkes2019compositionality,keysers2019measuring}. Identifiability results characterize assumptions for recovering latent structure from observations~\citep{locatello2018challenging,khemakhem2019vae}. Our controlled study evaluates whether complementary observations improve response prediction and whether learned dynamics transfer to new action orders.

\section{Shared Physical Representations Through Responses}
\label{sec:representation}

\subsection{Response Coordinates and Prediction}

We summarize each entity's responses to fitting queries by a low-dimensional coordinate. A sensor-specific compiler estimates this coordinate from observations, and a shared response model uses it to predict the response to a requested query.

We describe an entity by its responses under a specified apparatus and query family. Let $\omega\in\Omega$ denote an entity, $\mathcal A$ the set of basic interventions, and $q\in\mathcal A^*$ a finite intervention program. An apparatus $\alpha$ fixes the response sensor, contact geometry, initial condition, and preprocessing. The response law is
\begin{equation}
\Rop_{\omega}^{\alpha}(q)
:=\mathcal L\!\left(Y_{0:T(q)}\mid\operatorname{do}(q),\omega,\alpha\right).
\label{eq:response-operator}
\end{equation}
Here $\mathcal L$ denotes the response distribution, and $Y_{0:T(q)}$ the observations produced by $q$. In Cluster, $q$ specifies a scan condition; in the controlled system, it specifies a sequence of force pulses. We represent these responses by preprocessed spectra or trajectory vectors. A response chart assigns coordinates $\theta$ to training-entity response profiles, and a response model $G$ maps a coordinate and query to a prediction. We call $G$ an executor because it produces the response to the requested intervention.

Given observations $e_{m,P}$ from modality $m$ on evidence panel $P$, a compiler $C_{m,P}$ predicts the coordinate:
\begin{equation}
\widehat R_{m,P}(q)=G\!\left(C_{m,P}(e_{m,P}),q\right).
\label{eq:prediction}
\end{equation}
Sharing means that the sensor-specific estimates are interpreted through the same response chart and executor. We assess this relationship through their predictions, rather than requiring equality of the inferred coordinates. The coordinates describe response variation and need not equal physical parameters such as stiffness or damping. The executor receives neither modality nor panel identifiers. Appendix~\ref{app:formal} gives the associated response-equivalence relation.

\subsection{Cross-Sensor Consistency and Entity Specificity}
\label{sec:substitution}

Response substitution changes the sensor and evidence panel while keeping the entity and prediction queries fixed. We construct the chart and executor from training entities and freeze them before fitting each compiler separately. Two modalities $a$ and $b$ observe disjoint panels $P_0$ and $P_1$. Their predictions are compared in both orientations, $(a,P_0;b,P_1)$ and $(a,P_1;b,P_0)$, over queries $\Qbridge$ outside both panels. Thus agreement cannot be explained by presenting the same query instances to both branches.

For stacked response predictions $R,R'$, we use a squared distance normalized by training-entity variation:
\begin{equation*}
D_{\Qbridge}(R,R')=\frac{\|R-R'\|_F^2}{Z},\qquad
Z=\frac{1}{N_{\mathrm{tr}}}\sum_{j=1}^{N_{\mathrm{tr}}}
\|G_{\Qbridge}(\theta_j)-G_{\Qbridge}(0)\|_F^2.
\end{equation*}
Here $G_{\Qbridge}(0)$ is the population chart-center prediction. We average the two cross-panel distances over matching entities to obtain $D_{\mathrm{same}}$, and over different entities to obtain $D_{\mathrm{wrong}}$. The average distance from each branch to the population prediction is $D_{\mathrm{pop}}$. The response-substitution criterion is
\begin{equation}
D_{\mathrm{same}}<\min\{D_{\mathrm{wrong}},D_{\mathrm{pop}}\}.
\label{eq:response-substitution}
\end{equation}
The two controls distinguish entity-specific agreement from agreement near a common center. Prediction error against withheld targets supplies the separate accuracy measure, normalized by population prediction error. Appendix~\ref{app:cluster} specifies the finite estimators and evidence allocation.

\subsection{Complementary Information and New Action Orders}

For complementary observations, each compiler represents uncertainty over response coordinates with a probability distribution. Combining these beliefs can reduce uncertainty and improve response prediction. We evaluate this fusion benefit separately from cross-sensor substitution: complementary observations need not be individually sufficient to recover the same response function. Even when observations identify an entity's response coordinates, they may leave the effect of a new action order undetermined.

\begin{proposition}[Ambiguity of unseen action orders]
\label{prop:nonimplication}
Low-rank response compression on observed queries, strict squared-loss Bayes-risk reduction from each modality, and strict posterior-trace reduction under fusion do not suffice to identify an unseen ordered query.
\end{proposition}

Let observed primitive responses be $\Rop(A)=U$ and $\Rop(B)=V$, with independent Gaussian coordinates. Two modalities measure $U$ and $V$ separately with additive Gaussian noise. The observed responses have an exact two-dimensional chart; each measurement reduces Bayes risk, and fusion reduces posterior trace. Two systems can agree on these observations while assigning opposite composition laws:
\begin{equation*}
\Rop_{\sigma}(AB)=U+V+\sigma\gamma UV,\qquad
\Rop_{\sigma}(BA)=U+V-\sigma\gamma UV,
\quad\sigma\in\{-1,+1\},\ \gamma>0.
\end{equation*}
The systems induce identical evidence and primitive-response distributions but opposite order effects. A learner using the common observed law produces the same prediction for both systems and must err on at least one. Appendix~\ref{app:proof} gives the risk calculation. The observations leave the composition rule undetermined even if the entity coordinates $(U,V)$ are known. The controlled experiment examines whether reusing a learned state-update rule supports prediction under a new action order.

\section{Shared Physical Responses Across Sensors}
\label{sec:sensors}

\begin{figure}[t]
    \centering
    \includegraphics[width=\linewidth]{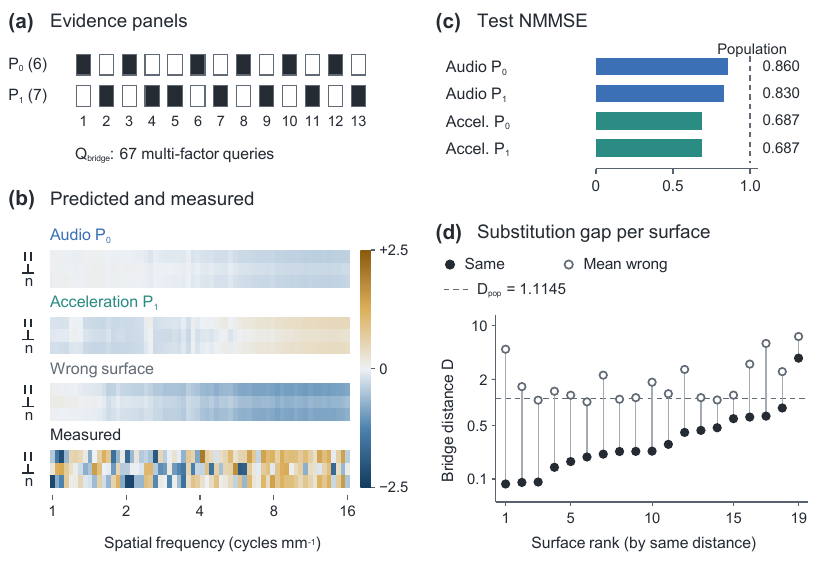}
    \caption{\textbf{Surface-specific responses inferred from audio and acceleration.}
(a) The 13 axis queries are split into disjoint evidence panels; the 67 remaining
queries form $\Qbridge$.
(b) Predicted and measured standardized log spatial spectra for surface 20 at
$(180^\circ,20\,\mathrm{mm/s},500\,\mathrm{mN})$, with wrong surface 56.
Measured responses use repeat 1. Rows denote parallel ($\parallel$), perpendicular
($\perp$), and normal (n) acceleration. All four groups share the color range $[-2.5,2.5]$ and a
logarithmic frequency axis. Predictions are constrained by the frozen low-rank response
chart; measured targets are shown without chart compression.
(c) Test NMMSE on the same 19 surfaces, with population error normalized to one.
(d) Same-surface and mean wrong-surface distances for 19 test surfaces, sorted by
same-surface distance. Thin lines connect paired values. Same-surface distance is lower for all 19 surfaces.
The population response-distance reference $D_{\mathrm{pop}}=1.1145$ uses a
different normalization from the NMMSE baseline.}
    \label{fig:cluster}
\end{figure}

\subsection{Surface Observations and Test Responses}

Cluster Haptic records synchronized acceleration, audio, force, and position while a controlled three-axis machine scans 118 surfaces under varied direction, velocity, normal force, and repeat conditions~\citep{eguchi2026cluster,eguchi2026clusterdata}. We split surface identities into 80 training, 19 development, and 19 test surfaces. The response target is a three-axis log spatial acceleration spectrum. Audio and acceleration provide different observations from which to infer this common target response.

Of 80 scan queries, 13 contain the reference condition or a change in one scan coordinate. Their fixed allocation gives disjoint evidence panels $P_0$ and $P_1$ with six and seven queries (Figure~\ref{fig:cluster}a). The remaining 67 multi-factor queries form $\Qbridge$. A velocity change alone is an axis query, for example, whereas a joint velocity and normal-force change belongs to the bridge set. Compilers receive repeat-0 evidence; repeat-1 measurements supply response targets. Test bridge responses are withheld until the chart, compilers, and panel assignment are fixed.

A training-only rank-four basis compresses each scan, and a rank-three SVD represents surface profiles concatenated across all 80 queries. These fixed linear maps form the shared chart and executor. Four linear-kernel compilers map audio or acceleration from either evidence panel to this chart. The training surfaces cover every query, so the experiment tests recovery of a new surface's responses from limited observations. The bridge queries are absent from the test surface's evidence, but their scan conditions are known to the response model. Both cross-panel orientations contribute equally to evaluation; Appendix~\ref{app:cluster} gives signal processing, exact estimators, and development results.

\subsection{Cross-Sensor Consistency and Predictive Accuracy}

Figure~\ref{fig:cluster}b compares predictions from audio and acceleration with a wrong-surface control and the measured response for one test surface and bridge query. All four branches predict measured responses better than the population baseline (Figure~\ref{fig:cluster}c). Test NMMSE is $0.8602$ and $0.8298$ for audio $P_0$ and $P_1$, and $0.6873$ and $0.6870$ for acceleration. Acceleration is more accurate under both panel assignments. Appendix~\ref{app:cluster} reports the corresponding development results.

The cross-sensor predictions also preserve surface identity. Mean same-surface distance is $D_{\mathrm{same}}=0.5161$, compared with $D_{\mathrm{wrong}}=2.3104$ and $D_{\mathrm{pop}}=1.1145$. The overall mean wrong-surface distance is approximately $4.5$ times the same-surface distance. Each of the 19 test surfaces has a smaller same-surface distance than its mean wrong-surface distance, with a median per-surface ratio of $5.2$ (Figure~\ref{fig:cluster}d). Cross-sensor consistency preserves surface-specific behavior while allowing the modalities to differ in predictive accuracy. Section~\ref{sec:orders} uses a separate controlled system to examine how response coordinates support prediction when familiar actions occur in a new order.

\section{Generalization Across Action Orders}
\label{sec:orders}

\subsection{Controlled Dynamics and Complementary Observations}
\label{sec:controlled-setup}

Cluster tests inference of a new entity within a response family whose scan conditions were covered during training. The controlled study asks whether a supplied entity coordinate supports prediction when familiar actions form new transitions. The governing equations stay fixed, so this tests reuse of learned dynamics across action orders. We use an elastoplastic system whose stiffness $k$, damping $c$, and yield force $F_y$ vary across entities. For unit mass and plastic displacement $p$, restoring force and motion satisfy
\begin{equation}
f=\operatorname{clip}(k(x-p),-F_y,F_y),\qquad
\dot v=u-cv-f,\qquad \dot x=v.
\label{eq:oscillator}
\end{equation}
The simulator updates plastic displacement after each velocity-first Euler step; Appendix~\ref{app:simulator} gives the exact recursion and initial conditions. Free decay starts from a small release that remains elastic, so its evidence is exactly invariant to $F_y$. Hysteresis records fully settled quasistatic responses, for which velocity is zero and damping has no effect. Both observations retain stiffness information. Training, development, and test entities occupy disjoint Cartesian grids of sizes $5^3$, $4^3$, and $4^3$.

Three force pulses $A$, $B$, and $C$ form finite programs. The 17 fitting programs contain all three primitives in all positions but only self transitions and $B\leftrightarrow C$. Development introduces $A\leftrightarrow C$. Test programs are $AB$ and $BA$; neither adjacency appears in fitting or development, including in longer programs. After each program, we simulate 60 additional steps with zero input and record displacement, velocity, and restoring force. The driven segment is excluded from scoring. Plastic memory makes the post-action response depend on which pulse occurs first, measured by the response difference
\begin{equation}
\Delta_{\omega}^{A,B}=r_{\omega}(AB)-r_{\omega}(BA).
\label{eq:commutator}
\end{equation}
Here $r_{\omega}(q)$ is the normalized post-action trajectory, and $\Delta$ is the response commutator. Its RMS across test entities, time, and channels is $0.62016$. Both parameters hidden from the single modalities affect this post-action response, providing a setting in which complementary observations can improve prediction.

A rank-three chart retains $98.95\%$ of centered fitting-response variation. To assess execution separately from sensory inference, an oracle supplies an entity's coordinate from its fitting-query response profile. For the executor comparison in Section~\ref{sec:executor-comparison}, NMSE divides prediction MSE by the error of a common query-only predictor within each split. This evaluation asks whether the dynamics model can use the response coordinate before introducing uncertainty from sensor observations.

\begin{figure}[t]
    \centering
    \includegraphics[width=\linewidth]{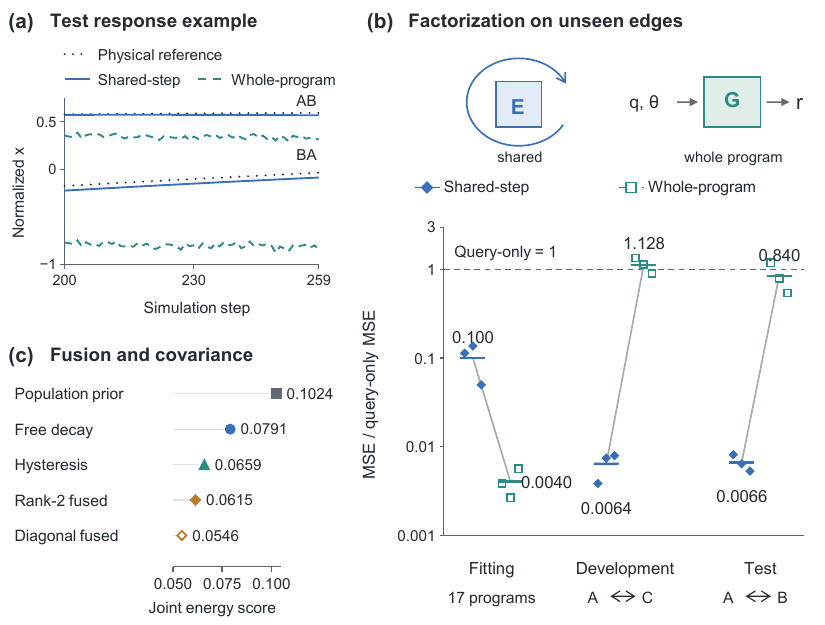}
    \caption{\textbf{Action-order generalization and complementary fusion.}
(a) Simulator displacement and oracle-coordinate predictions for test entity 32, seed 0, on both action orders. The entity index was fixed before training; seed 0 is the first initialization. All curves share training-based normalization.
(b) Three independent initializations per executor at a 30,000-update budget. Small points show individual runs; horizontal bars and labels show their means. Both executors use the same query-only baseline within each split. Checkpoints are selected by development post-action response MSE. The example in (a) uses these selected models.
(c) Joint energy score for fusion through a separately trained frozen executor, with lower values better.
The diagonal compiler is trained separately from the rank-two-factor compiler.
}
    \label{fig:ordered}
\end{figure}

\subsection{Reusing Step Dynamics on Unseen Transitions}
\label{sec:executor-comparison}

This diagnostic compares two executors using the same response coordinate. The shared-step, or step-factorized, model applies a residual MLP repeatedly,
\begin{equation*}
s_{t+1}=s_t+E_{\phi}(s_t,u_t/4,\theta),
\end{equation*}
using normalized observables $s_t=(x_t,v_t,f_t)$, the rescaled force input, and the coordinate. Two hidden layers of 128 SiLU units map these seven inputs to three state increments, giving 17,923 trainable parameters. The coordinate stays fixed throughout a program, while the state carries its action history.

Let $\mathcal T_{\mathrm{fit}}$ denote the empirical distribution of simulator transitions from fitting entities and programs. Training minimizes
\begin{equation}
\mathcal L_{\mathrm{step}}(\phi)=
\mathbb E_{(s,u,s',\theta)\sim\mathcal T_{\mathrm{fit}}}
\left[\frac{1}{3}\left\|E_{\phi}(s,u/4,\theta)-(s'-s)\right\|_2^2\right].
\label{eq:step-loss}
\end{equation}
Minibatches sample 128 transitions uniformly with replacement. The targets include intermediate states during the driven segment as well as the post-action response. Thus the shared-step executor receives local transition supervision throughout each fitting trajectory. At evaluation, it starts from the normalized zero physical state and recursively uses its own predictions for the full force sequence and post-action response. No measured intermediate state is supplied during this rollout. 

The whole-program executor embeds each action token into 32 dimensions and encodes the ordered word with a 32-dimensional GRU. A $35\rightarrow52\rightarrow180$ SiLU head combines its final hidden state with the three response coordinates to predict the $60\times3$ post-action trajectory. This model has 17,844 parameters and minimizes post-action response MSE over batches of 128 entity--program pairs. Both executors use the same chart and fitting programs; the shared-step model receives intermediate transition targets, while the whole-program model receives post-action trajectory targets.

Figure~\ref{fig:ordered}a shows both models predicting the same test entity. The shared-step predictions follow the simulator more closely; the whole-program predictor underestimates displacement under both orders.

For the architecture comparison in Figure~\ref{fig:ordered}b, we train each executor from three independent initializations for 30,000 updates. Both use AdamW with learning rate $10^{-3}$ and weight decay $10^{-5}$, selecting between 15,000 and 30,000 updates by development post-action response MSE. A separately trained query-only GRU supplies one common denominator per split. It receives the action word without an entity coordinate and predicts the complete post-action response. All checkpoint selections are finalized before test responses are evaluated. Appendix~\ref{app:executor-comparison} specifies this comparison and reports each run.

The whole-program executor fits the 17 observed programs more accurately, with mean NMSE $0.00404$ compared with $0.09990$ for shared-step execution. The ranking reverses on both unseen edges in every initialization. Development NMSE is $1.128$ versus $0.00641$; test NMSE is $0.840$ versus $0.00663$, in the same model order. Test MSE before division by the common baseline is $0.76948$ versus $0.00607$. The whole-program mean improves on the query-only test baseline, but its errors remain much larger than those of shared-step execution. This reversal therefore holds for the original MSE as well as the normalized comparison.

Shared step dynamics carry the effect of the first action into the execution of the second through the evolving state. In $AB$, pulse $B$ acts on the displacement, velocity, and restoring force left by $A$; reversing the order changes that state. The same update rule processes both sequences. Reusing this rule provides a mechanism for applying learned dynamics to new action orders. The comparison tests this recursive structure together with its local supervision; matching the chart and update budget does not separate those two design choices.

\subsection{Combining Complementary Evidence}

For the fusion analysis, we freeze a separately trained shared-step executor selected by development oracle error under a 30,000-update budget. Two 32-dimensional GRU compilers map free-decay and hysteresis observations to Gaussian information factors over the response coordinates (Figure~\ref{fig:overview}b). Appendix~\ref{app:orq} reports the executor's training-budget analysis.

A standard-normal prior combines these factors by adding their natural parameters under the Gaussian product-of-experts rule~\citep{hinton2002products,wu2018multimodal,sutter2021generalized}:
\begin{equation}
\Sigma_S=(I+\textstyle\sum_{m\in S}\Lambda_m)^{-1},\qquad
\mu_S=\Sigma_S\textstyle\sum_{m\in S}\eta_m.
\label{eq:info-fusion}
\end{equation}
Each factor uses $\Lambda_m=L_mL_m^\top$ with $L_m\in\mathbb R^{3\times2}$. Both compilers are trained jointly on true fitting responses through the frozen executor, with single-modality and fusion losses.

Joint energy score evaluates the predicted $AB/BA$ response distribution~\citep{gneiting2007proper}. Each coordinate sample is executed under both orders, yielding a joint vector of $2\times60\times3=360$ response values. The same sampled entity therefore links the two predicted trajectories. Mean-response NMSE evaluates their predictive means relative to the executed population prior. Order-effect error compares the difference of those means with the true $AB-BA$ response, using the true order-effect RMS as its scale. A zero predicted order effect has error one. These definitions retain both response accuracy and the difference induced by changing the action order (Appendix~\ref{app:orq}).

Combining observations improves prediction over either source alone (Figure~\ref{fig:ordered}c; Table~\ref{tab:orq}a). Response NMSE decreases from $0.5492$ for free decay and $0.4066$ for hysteresis to $0.3193$ after fusion; joint energy score also improves. Replacing evidence with an observably different sequence from another entity matched in stiffness raises NMSE to $0.8398$. This control supports the contribution of entity-specific complementary information to the fusion improvement.

The controlled blind spots make this gain physically interpretable. Free decay supplies damping information that settled hysteresis lacks, while hysteresis supplies yield information absent from the elastic release. Both quantities affect the post-action response. Fusion reduces posterior trace by $68\%$ relative to the more informative single modality. The information sum is full rank, with minimum eigenvalue $0.5867$, so the learned factors jointly constrain all three chart coordinates.

The separately trained diagonal model gives the lowest errors on all three reported metrics, including order-effect prediction (Table~\ref{tab:orq}b). Both fused distributions also outperform the point-valued control on energy score and NMSE. The diagonal factors allow rank three, whereas the model with off-diagonal information uses rank-two factors. This comparison favors the diagonal formulation while changing both the per-modality rank constraint and the allowed off-diagonal structure.

\begin{table}[t]
\caption{\textbf{Fusion and belief parameterization for test $AB/BA$ responses.}
Results on 64 test entities with the frozen executor used for fusion.
NMSE is relative to the executed population prior; order-effect error is relative
RMS error. Lower is better. Single-modality order-effect errors were not reported.
Diagnostic controls are reported in Appendix~\ref{app:scores}.}
\label{tab:orq}
\centering
\small
\renewcommand{\arraystretch}{1.08}
\setlength{\tabcolsep}{7pt}
\begin{tabularx}{\linewidth}{@{}Xrr@{}}
\toprule
\multicolumn{3}{@{}l}{\textbf{(a) Response prediction}}\\
\addlinespace[3pt]
Model & Joint ES $\downarrow$ & NMSE $\downarrow$\\
\midrule
Population prior & 0.10239 & 1.0000\\
\addlinespace[3pt]
Free decay & 0.07909 & 0.5492\\
Hysteresis & 0.06594 & 0.4066\\
\addlinespace[3pt]
Rank-2 fused & 0.06147 & 0.3193\\
Diagonal fused & \textbf{0.05456} & \textbf{0.2873}\\
Point fused & 0.08484 & 0.3736\\
\bottomrule
\end{tabularx}

\vspace{8pt}
\begin{tabularx}{\linewidth}{@{}Xr@{}}
\toprule
\multicolumn{2}{@{}l}{\textbf{(b) Order-effect prediction}}\\
\addlinespace[3pt]
Model & Relative RMS error $\downarrow$\\
\midrule
Population prior & 0.37388\\
Rank-2 fused & 0.18165\\
Diagonal fused & \textbf{0.17396}\\
Point fused & 0.20158\\
\bottomrule
\end{tabularx}
\end{table}

\FloatBarrier
\section{Discussion and Limitations}

In Cluster, sharing means recovering surface-specific acceleration responses from either sensory source. This evaluation is asymmetric: acceleration evidence shares its sensing modality with the response target, while audio evidence does not. This may contribute to acceleration's lower prediction error. Using audio responses as the target would test whether the same-versus-wrong distance gap persists under the reciprocal choice of response apparatus.

Selecting an executor by fitting error would favor the whole-program predictor, whereas evaluation on new action transitions favors shared-step execution. The shared-step model carries action history through its recursively updated state. A targeted follow-up would train the same model with only a post-action rollout loss, testing how much its advantage depends on local transition supervision.

Fusion determines the entity information available to the executor. The exact blind spots create a reason to combine observations, and fused beliefs improve response prediction over either single modality. The fused model also predicts the order effect more accurately than the population predictor. The diagonal formulation gives the best result among the tested belief models. Together with Proposition~\ref{prop:nonimplication}, these comparisons separate the complementary observations that constrain an entity from the execution rule that maps its coordinates and an action sequence to a response.

The present evidence covers one real apparatus, two modalities, and 19 test surfaces, together with a controlled system evaluated on 64 test entities. Cluster tests new entities within a known query grid; the controlled system additionally withholds action transitions. Extending this design to real sequential contact tasks would connect the controlled analysis with richer interaction dynamics.

\section{Conclusion}

We evaluated shared physical representations through responses predicted from different sensory observations. On Cluster Haptic, separately fitted audio and acceleration compilers preserve surface-specific behavior and improve on population prediction. In a separate controlled system, shared step dynamics predict new action orders more accurately than a whole-program predictor across three independent initializations. Complementary fusion improves prediction, with the diagonal belief performing best among the evaluated fusion models.

\clearpage

\subsection*{AI use statement}

In this work, we used generative AI tools to support research ideation and execution, retrieve and identify relevant literature, and edit the manuscript for clarity and readability. We reviewed all AI-assisted work and take responsibility for the final content of this work, including text, claims, and artifacts produced with the aid of generative AI.

\bibliography{iclr2027_conference}
\bibliographystyle{iclr2027_conference}

\appendix

\section{Formal Definitions and Ambiguity Construction}
\label{app:formal}

\subsection{Response Equivalence}

For a fixed apparatus $\alpha$ and query family $\mathcal{Q}$, define
\begin{equation*}
\omega\equiv_{\alpha,\mathcal{Q}}\omega'
\quad\Longleftrightarrow\quad
\Rop_{\omega}^{\alpha}(q)=\Rop_{\omega'}^{\alpha}(q)
\quad\text{for every }q\in\mathcal{Q}.
\end{equation*}
The quotient $\Omega/{\equiv_{\alpha,\mathcal{Q}}}$ groups entities with the same response law under these observations. A response chart gives coordinates to the response profiles represented by the training data. Its scope is therefore conditional on both the apparatus and the query family. For a belief $\nu$ over coordinates, the predictive law over the stacked responses is the pushforward $(G_{\mathcal{Q}})_{\#}\nu$.

\subsection{Proof of Proposition~\ref{prop:nonimplication}}
\label{app:proof}

Let $U,V\overset{\mathrm{iid}}{\sim}\mathcal{N}(0,1)$ and let the two modalities observe
\begin{equation}
E_1=U+\epsilon_1,\qquad E_2=V+\epsilon_2,\qquad
\epsilon_1,\epsilon_2\overset{\mathrm{iid}}{\sim}\mathcal{N}(0,\tau^2),\qquad 0<\tau^2<\infty.
\end{equation}
For observed primitive queries, set $\Rop(A)=U$ and $\Rop(B)=V$. The response profile lies exactly in the rank-two chart $Z=(U,V)$. Writing $\alpha=\tau^2/(1+\tau^2)$, the unimodal and fused posterior covariances are
\begin{equation}
\Sigma_1=\operatorname{diag}(\alpha,1),\quad
\Sigma_2=\operatorname{diag}(1,\alpha),\quad
\Sigma_{12}=\alpha I.
\end{equation}
Thus $\operatorname{tr}\Sigma_{12}=2\alpha<1+\alpha=\operatorname{tr}\Sigma_1=\operatorname{tr}\Sigma_2$.

Now index two systems by an unobserved composition law $\sigma\in\{-1,+1\}$:
\begin{align}
\Rop_{\sigma}(AB)&=U+V+\sigma\gamma UV,\\
\Rop_{\sigma}(BA)&=U+V-\sigma\gamma UV,\qquad \gamma>0.
\end{align}
The systems induce the same joint distribution over all modality evidence and observed-query responses. Both ordered responses have population mean zero. Under squared loss, each modality strictly reduces Bayes risk for either ordered response because, for example,
\begin{equation}
\mathbb{E}[\Rop_{\sigma}(AB)\mid E_1]=\frac{E_1}{1+\tau^2},
\end{equation}
whose variance $1/(1+\tau^2)$ is the Bayes-risk reduction relative to the zero population predictor.

The commutator is $C_{\sigma}=2\sigma\gamma UV$. Any learner using only the common observed law produces the same predictor $\widehat C$ under both signs. Pointwise,
\begin{equation}
\frac{(\widehat C-2\gamma UV)^2+(\widehat C+2\gamma UV)^2}{2}
=\widehat C^2+4\gamma^2U^2V^2.
\end{equation}
Taking expectations and using $\mathbb{E}[U^2V^2]=1$ shows that at least one indistinguishable system has commutator MSE at least $4\gamma^2$. Exact low-rank compression, useful marginal evidence, and reduced fused uncertainty therefore leave the unseen composition law unidentified. \hfill$\square$

\section{Cluster Haptic Implementation and Additional Results}
\label{app:cluster}

We use Cluster Haptic v5 with a category-balanced split of 80 training, 19 development, and 19 test surface identities. All response normalizers, chart bases, and compiler statistics use the training identities.

\paragraph{Response construction.}
We rotate acceleration into path-parallel, path-perpendicular, and surface-normal axes. The central two-thirds of the measured scan path is resampled at 5~kHz. Welch spectra use 20-mm Hann windows with 10-mm hops in spatial coordinates. Each axis is interpolated to 64 frequencies from 1 to 16 cycles/mm, producing a $3\times64$ natural-log spatial-PSD response on a geometric frequency grid. Audio evidence contains 32 $\log_{10}$ mean-PSD values over frequency bands from 100~Hz to 18~kHz.

\paragraph{Frozen chart and compilers.}
A rank-four basis compresses each standardized 192-dimensional acceleration response. Concatenating coefficients across all 80 queries gives a 320-dimensional surface profile; its centered, fixed-sign rank-three SVD defines $\theta\in\mathbb{R}^3$. For branch $(m,P)$, the compiler uses standardized evidence dimension $d_{m,P}$ and the fixed kernel
\begin{equation*}
K_{m,P}(x,x')=1+\frac{x^{\top}x'}{d_{m,P}},\qquad
\widehat\theta=K_{x,T}(K_{T,T}+10^{-3}I)^{-1}\Theta_T.
\end{equation*}
The panel assignment is fixed before fitting. No surface identifier or material category enters a compiler.

\paragraph{Evidence and target separation.}
The safe cache contains repeat-0 axis evidence for all surfaces, training repeat-1 responses for all 80 queries, and development repeat-1 responses for the 67 bridge queries. Test repeat-1 responses are absent. The 13 axis queries are assigned to $P_0$ or $P_1$ by the fixed parity of their categorical query-coordinate codes, yielding six and seven queries. This allocation is fixed before fitting the four compilers. No surface identifier or material category is supplied as an input, although the branches share response-coordinate targets for the same training surfaces.

\paragraph{Response-distance estimators.}
For test entity $i$, write $\widehat R_u^{(i)}=[G(C_u(e_{i,u}),q)]_{q\in\Qbridge}$ for branch $u$. Let $\mathcal O=\{(a,P_0;b,P_1),(a,P_1;b,P_0)\}$ contain the two cross-panel orientations, $\mathcal U$ the four branches, and $\bar R=G_{\Qbridge}(0)$. Using the training-normalized squared distance of Section~\ref{sec:substitution},
\begin{align*}
D_{\mathrm{same}}&=\frac{1}{2N}\sum_{i=1}^{N}\sum_{(u,v)\in\mathcal O}
D_{\Qbridge}(\widehat R_u^{(i)},\widehat R_v^{(i)}),\\
D_{\mathrm{wrong}}&=\frac{1}{2N(N-1)}\sum_{i\ne j}\sum_{(u,v)\in\mathcal O}
D_{\Qbridge}(\widehat R_u^{(i)},\widehat R_v^{(j)}),\\
D_{\mathrm{pop}}&=\frac{1}{4N}\sum_{i=1}^{N}\sum_{u\in\mathcal U}
D_{\Qbridge}(\widehat R_u^{(i)},\bar R).
\end{align*}
For $N=19$, these average 38 same-entity, 684 wrong-entity, and 76 population terms. The per-surface wrong distance in Figure~\ref{fig:cluster}d first averages over the other 18 surfaces and then over the two orientations. Each surface's same distance is smaller than this mean wrong distance. This ordering does not require every individual wrong pairing to be farther away.

\paragraph{Prediction error and development results.}
NMMSE is the sum of squared errors across the standardized response features, queries, and entities divided by the corresponding population prediction error. Table~\ref{tab:cluster-dev} reports development and test values; Figure~\ref{fig:cluster}c shows the test comparison. This error normalization differs from the training-response variation used for $D_{\Qbridge}$.

\paragraph{Evaluation protocol.}
The split, query panels, normalizers, response chart, and compiler coefficients are fixed before accessing test repeat-1 targets. Those targets are subsequently extracted from the raw archive with the same feature pipeline to evaluate predictive accuracy. No chart, compiler, threshold, or panel assignment was revised after target access, and all four branches are reported.

\begin{table}[t]
\caption{\textbf{Response prediction on Cluster Haptic.} Repeat-1 NMMSE on 19 development and 19 test surfaces, each evaluated on 67 bridge queries. The population chart-center baseline equals one. Test targets enter no fitting or model selection.}
\label{tab:cluster-dev}
\centering
\small
\setlength{\tabcolsep}{12pt}
\begin{tabular}{@{}lrrrr@{}}
\toprule
& \multicolumn{2}{c}{Development} & \multicolumn{2}{c}{Test}\\
\cmidrule(lr){2-3}\cmidrule(l){4-5}
Modality & $P_0$ & $P_1$ & $P_0$ & $P_1$\\
\midrule
Audio & 0.8636 & 0.8050 & 0.8602 & 0.8298\\
Acceleration & 0.7328 & 0.7386 & 0.6873 & 0.6870\\
Population & 1.0000 & 1.0000 & 1.0000 & 1.0000\\
\bottomrule
\end{tabular}
\end{table}

\section{Controlled-System Implementation and Additional Results}
\label{app:orq}

\subsection{Simulator State and Discrete Updates}
\label{app:simulator}

The simulator stores displacement $x_t$, velocity $v_t$, plastic displacement $p_t$, and restoring force $f_t$. With unit mass and $h=\Delta t=0.005$, one force-input step applies
\begin{align}
v_{t+1}&=v_t+h(u_t-cv_t-f_t), &
x_{t+1}&=x_t+h v_{t+1},\label{eq:discrete-motion}\\
f^{\mathrm{trial}}_{t+1}&=k(x_{t+1}-p_t), &
f_{t+1}&=\operatorname{clip}(f^{\mathrm{trial}}_{t+1},-F_y,F_y),\label{eq:force-return}\\
p_{t+1}&=\begin{cases}
x_{t+1}-f_{t+1}/k,& |f^{\mathrm{trial}}_{t+1}|>F_y,\\
p_t,& |f^{\mathrm{trial}}_{t+1}|\leq F_y.
\end{cases}\label{eq:plastic-update}
\end{align}
Velocity is updated before displacement, using the previous restoring force. The force and plastic state are then updated at the new displacement. During elastic loading or unloading, $p_t$ remains fixed; after either yield threshold is exceeded, the plastic update keeps the restoring force at that threshold. These are the steps implemented by \texttt{\_orq\_simulate\_bank} and \texttt{\_orq\_update\_force} in \texttt{pilot\_cross\_context.py}.

Each action program starts at $x_0=v_0=p_0=f_0=0$. The stored trajectory contains $(x_{t+1},v_{t+1},f_{t+1})$ after each input step, and the final 60 stored states form the scored response. Free decay instead starts at $x_0=0.035$, $v_0=p_0=0$, and $f_0=kx_0$, with $u_t=0$ throughout. Quasistatic hysteresis starts with $p=0$ and imposes a 90-point displacement path $0\rightarrow0.55\rightarrow-0.55\rightarrow0.35\rightarrow0$. Its four segments contain 17, 33, 27, and 13 equally spaced points, respectively; each excludes its right endpoint except the last. At each imposed displacement, the same force-return and plastic updates apply with velocity fixed to zero. No dynamic integration is used for this settled observation.

\subsection{Entity and Query Splits}

The simulator uses $\Delta t=0.005$. Training entities form a $5^3$ grid over $k\in[1.2,2.4]$, $c\in[0.12,0.52]$, and $F_y\in[0.18,0.42]$. Development and test entities occupy disjoint interleaved $4^3$ grids. The free-decay modality records 200 steps from $x_0=0.035$; its maximum elastic force remains below the minimum yield force. The hysteresis modality records a 90-point fully settled quasistatic path.

Primitive pulses have amplitudes $u_A=4.0$, $u_B=-2.8$, and $u_C=1.4$, each with 70 driven and 30 coast steps. The action-program split is
\begin{align*}
\Qfit={}&\{A,B,C,AA,BB,BC,CB,CC,AAA,BBB,BBC,BCB,\\
&BCC,CBB,CBC,CCB,CCC\},\\
\Qdev={}&\{AC,CA,AAC,ACA,ACB,ACC,BCA,CAA,CAC,CCA\},\\
\Qtest={}&\{AB,BA\}.
\end{align*}
Length-three programs containing an $AB$ or $BA$ adjacency are excluded from fitting and development. The 17 fitting, 10 development, and two test programs, together with these 10 excluded programs, partition all 39 words of length one to three.

\subsection{Exact Blindness and Response Relevance}

For free decay, the initial displacement satisfies $\max_k k|x_0|<\min F_y$. Free decay can only reduce mechanical energy, so the trajectory remains elastic and $p=0$ throughout. Its complete evidence sequence is therefore invariant to yield force. Every hysteresis point is evaluated after the system has fully settled. Velocity is exactly zero, the damping term vanishes, and the hysteresis sequence is invariant to $c$.

The test response remains sensitive to every hidden coordinate. Adjacent-level changes in stiffness, damping, and yield force produce normalized response RMS values $0.09509$, $0.03925$, and $0.09041$. The $AB/BA$ commutator RMS is $0.62016$. Wrong-entity fusion is matched within stiffness and replaces evidence only when at least one complementary coordinate changes. The wrong-entity control includes 1,536 valid fused-entity terms and 3,072 response terms; every replacement changes observable evidence in at least one complementary coordinate.

\subsection{Response Chart, Executor, and Compilers}

Each fitting entity contributes a flattened normalized profile of size $17\times60\times3=3060$. The chart is centered before SVD; each component sign is fixed by requiring the largest-magnitude loading to be positive; scores are divided by their train standard deviations. The relative Frobenius residual is
\begin{equation}
\sqrt{\frac{\|R-\widehat R\|_F^2}{\|R-\bar R\|_F^2}}=0.102265,
\end{equation}
equivalent to unexplained energy $0.0104582$.

The step executor receives normalized $(x,v,f)$, the action scaled by $1/4$, and three chart coordinates. Its residual MLP has widths $7\rightarrow128\rightarrow128\rightarrow3$, SiLU activations, and 17,923 parameters. It receives neither plastic displacement $p$ nor the true parameter vector $(k,c,F_y)$. Training minimizes one-step increment MSE using simulator states and transitions from the complete fitting trajectories, including their driven portions. AdamW uses learning rate $10^{-3}$, weight decay $10^{-5}$, and batch size 128. At test time, the same network is applied recursively without observed intermediate states.

\begin{figure}[t]
\centering
\includegraphics[width=176pt]{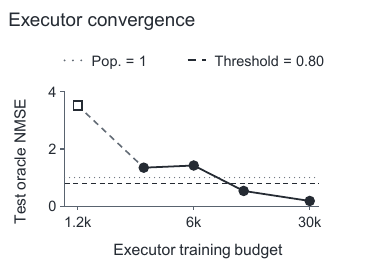}
\caption{\textbf{Executor training-budget analysis.} Each point reports the checkpoint selected by development oracle NMSE within the stated budget. The initial 1,200-update model does not meet the predefined $0.80$ threshold. Larger budgets are evaluated in a separate convergence analysis. Only reported test values are connected, including the increase at 6,000 updates. The denominator is the corresponding executor's zero-coordinate error.}
\label{fig:budget}
\end{figure}

The initial update budget is 1,200. Development oracle NMSE selects between checkpoints at half and all of the budget, hence 600 and 1,200 updates for this run. The same selection rule is used for extended budgets of 3,000, 6,000, 12,000, and 30,000 updates. Figure~\ref{fig:budget} reports these budgets; the retained result series does not separately record the selected update count at each budget. The runner increases compiler budgets along with executor budgets and refits the compilers for each analysis. The splits, chart construction, step architecture, seed, and thresholds stay fixed. Oracle metrics bypass all compilers and therefore diagnose the executor.

For budgets of 1,200, 3,000, 6,000, 12,000, and 30,000 updates, test oracle NMSE is $3.5175$, $1.3464$, $1.4220$, $0.5348$, and $0.1844$, respectively. Development oracle NMSE is $1.2238$ at the initial budget and $0.0756$ at the final budget.

Each modality encoder is a 32-dimensional GRU. In the rank-two-factor model, its head returns $L_m\in\mathbb R^{3\times2}$ and $a_m\in\mathbb R^2$, with $\Lambda_m=L_mL_m^\top$ and $\eta_m=L_ma_m$. A separate diagonal model returns scales $l_m\in\mathbb R^3$ and observations $a_m\in\mathbb R^3$, with $\Lambda_m=\operatorname{diag}(l_m^2)$ and $\eta_m=l_m\odot a_m$; this permits rank three. Both are trained from their own initialization. The reported minimum fused-information eigenvalue refers to $\Lambda_a+\Lambda_b$, excluding prior precision $I$. It is distinct from the rank of either factor and from the covariance $(I+\Lambda_a+\Lambda_b)^{-1}$.

Compiler training compares the executed free-decay, hysteresis, and fused predictive distributions with true fitting responses, averaging their three energy scores. The executor remains frozen throughout this optimization. AdamW uses learning rate $10^{-3}$ and weight decay $10^{-5}$; each update samples eight entities and one fitting query. The point control has separate three-dimensional modality heads and a learned fusion head, and is trained with the corresponding singleton-distribution score.

The direct diagnostic encodes both evidence sequences with GRUs, encodes the ordered action tokens with another GRU, and predicts the $60\times3$ post-action trajectory with an MLP. It bypasses both chart and executor. It is distinct from the whole-program executor in Figure~\ref{fig:ordered}b, which receives the response coordinate and ordered action tokens.

\paragraph{Three-seed executor comparison.}
\label{app:executor-comparison}
We run this comparison separately from the budget and fusion analyses. The simulator, 125/64/64 entity grids, 17/10/2 program split, training-response normalizer, and rank-three chart remain fixed. Each architecture uses initialization seeds 0, 1, and 2, with a separate minibatch generator initialized from the corresponding seed. The shared-step network and transition loss are specified above. The whole-program network has a three-token embedding of width 32, a one-layer GRU with 32 hidden units, and a $35\rightarrow52\rightarrow180$ head with SiLU between its linear layers. Its input concatenates the final valid-token hidden state and the entity's three chart coordinates. Packing uses the true word lengths, so padding never acts as an extra primitive. The output is the normalized $60\times3$ post-action trajectory. The model has 17,844 trainable parameters, compared with 17,923 for the shared-step network.

Both architectures use AdamW with learning rate $10^{-3}$, weight decay $10^{-5}$, $(\beta_1,\beta_2)=(0.9,0.999)$, and $\epsilon=10^{-8}$. Each run makes 30,000 updates with batch size 128. Shared-step batches sample transitions; whole-program batches sample entity--program pairs. The development post-action response MSE selects between 15,000 and 30,000 updates for both models. All three shared-step runs select 30,000 updates; whole-program seeds 0, 1, and 2 select 15,000, 30,000, and 15,000 updates. No test entity response enters these choices.

The query-only baseline uses the same token embedding and GRU, with a $32\rightarrow52\rightarrow180$ head and 17,688 parameters. It receives no entity coordinate or sensory evidence. One run with seed 1001 uses the same optimizer, budget, and development selection rule, selecting 15,000 updates. At evaluation, its predicted post-action response for a query is identical for every entity. The same frozen predictions supply the denominator for all six executor runs, including for the unseen programs. Table~\ref{tab:executor-runs} reports MSE before this division. We save both candidate checkpoints, selected weights, all minibatch losses, per-entity predictions, targets, and normalizers in \texttt{outputs/executor\_comparison}; \texttt{executor\_comparison.py} reproduces the experiment.

\begin{table}[t]
\centering
\small
\caption{\textbf{Executor comparison before baseline normalization.} MSE averages over entities, programs, 60 post-action steps, and three channels standardized using training responses. Selection uses development MSE. The fixed query-only row supplies a common denominator within each column.}
\label{tab:executor-runs}
\begin{tabular}{lrrrrr}
\toprule
Model & Seed & Selected update & Fitting MSE & Dev. MSE & Test MSE \\
\midrule
Query-only & 1001 & 15,000 & 0.038488 & 0.202754 & 0.916450 \\
Shared-step & 0 & 30,000 & 0.004363 & 0.000780 & 0.007462 \\
Whole-program & 0 & 15,000 & 0.000147 & 0.273157 & 1.086167 \\
Shared-step & 1 & 30,000 & 0.005254 & 0.001508 & 0.005883 \\
Whole-program & 1 & 30,000 & 0.000102 & 0.230814 & 0.724599 \\
Shared-step & 2 & 30,000 & 0.001918 & 0.001610 & 0.004869 \\
Whole-program & 2 & 15,000 & 0.000217 & 0.181938 & 0.497677 \\
\bottomrule
\end{tabular}
\end{table}

\subsection{Predictive Scores and Normalization}
\label{app:scores}

For posterior $\mathcal{N}(\mu,\Sigma)$, a Cholesky factor maps the eight lexicographically ordered nodes in $\{-1,+1\}^3$ to $z_s=\mu+L_{\Sigma}s$, each with weight $w_s=1/8$. Executing each node on $AB$ and $BA$ gives a $D=2\times60\times3=360$ dimensional response vector. For predictive law $F$, the joint energy score is
\begin{equation}
\ES(F,r)=\frac{1}{\sqrt D}\left(\mathbb E\|X-r\|_2-\frac{1}{2}\mathbb E\|X-X'\|_2\right),
\quad X,X'\overset{\mathrm{iid}}{\sim}F.
\end{equation}
The implementation evaluates the finite-distribution form
\begin{equation}
\sum_s w_s\frac{\|\widehat r_s-r\|_2}{\sqrt{360}}
-\frac12\sum_{s,t}w_sw_t\frac{\|\widehat r_s-\widehat r_t\|_2}{\sqrt{360}}.
\end{equation}
The score is proper for this finite predictive distribution; the eight-node representation is not the exact nonlinear pushforward of a continuous Gaussian.

Three error normalizations serve different comparisons. For training budget $B$, let $G_B$ be the executor selected by development error and let $\operatorname{MSE}_{\mathcal E,\mathcal Q}$ average squared error over the evaluated entities, queries, time steps, and normalized channels. Figure~\ref{fig:budget} uses
\begin{equation}
\operatorname{OracleNMSE}(B)=
\frac{\operatorname{MSE}_{\mathcal E,\mathcal Q}[G_B(\theta_i^*,q),r_i(q)]}
{\operatorname{MSE}_{\mathcal E,\mathcal Q}[G_B(0,q),r_i(q)]}.
\end{equation}
Both errors use the same selected executor and evaluation set. Its zero-coordinate prediction changes with $B$. The runner emits \texttt{oracle\_response\_mse} and \texttt{population\_response\_mse}, but the saved budget series retains their ratio only. The raw errors and selected update counts cannot be recovered from these ratios.

Figure~\ref{fig:ordered}b instead uses the query-only predictor $H(q)$ defined above. For executor $G_{m,s}$ with architecture $m$ and seed $s$,
\begin{equation}
\operatorname{NMSE}_{m,s}(\mathcal E,\mathcal Q)=
\frac{\operatorname{MSE}_{\mathcal E,\mathcal Q}[G_{m,s}(\theta_i^*,q),r_i(q)]}
{\operatorname{MSE}_{\mathcal E,\mathcal Q}[H(q),r_i(q)]}.
\end{equation}
The denominator is $0.0384883$ on fitting programs, $0.2027545$ on development programs, and $0.9164496$ on test $AB/BA$. Within each split, it is identical for every architecture and seed. Figure~\ref{fig:ordered}b shows all three run values and their arithmetic mean. Because the denominator is shared, the relative ordering of MSE and NMSE is identical.

Table~\ref{tab:orq} uses mean-response MSE divided by the MSE of the executed population prior $\mathcal N(0,I)$. That prior is evaluated using the same eight nodes, so its mean prediction generally differs from $G(0,q)$ for a nonlinear executor. The oracle budget analysis, three-seed executor comparison, and fusion analysis consequently use distinct model selections and normalizations.

For commutator error, let $\widehat\Delta_i$ be the difference between predictive mean post-action responses for $AB$ and $BA$ on entity $i$. The relative RMS error is
\begin{equation}
\frac{\sqrt{\operatorname{mean}_{i,t,d}(\widehat\Delta_{i,t,d}-\Delta_{i,t,d})^2}}
{\sqrt{\operatorname{mean}_{i,t,d}\Delta_{i,t,d}^2}}.
\end{equation}
A zero predicted order effect therefore has relative error one. This metric is a relative RMS error, not an NMSE.

\paragraph{Diagnostic controls.}
Wrong-entity fusion has joint energy score $0.10217$ and NMSE $0.8398$ on the same test responses. Its order-effect error is unreported. The direct diagnostic, which bypasses the response chart and executor, has joint energy score $0.96737$, NMSE $44.873$, and order-effect error $0.47015$. Its architecture is described above; it is distinct from the whole-program executor in Figure~\ref{fig:ordered}b.

The zero-order-effect control predicts $\widehat\Delta=0$, giving relative order-effect error $1.00000$. It specifies only the difference between the two orders, so it does not define complete response predictions for computing energy score or NMSE. Free-decay and hysteresis order-effect errors were not reported, and their test predictive trajectories were not retained. Table~\ref{tab:orq}b therefore reports the four available population, fused, and point-belief comparisons.

\clearpage
\subsection{Additional Controlled-System Diagnostics}

The controlled system separates complementary sensory information from action-order prediction. The entity and program splits, physical parameters, response chart, and oracle threshold are fixed before evaluating test responses. Table~\ref{tab:controlled-diagnostics} summarizes the physical and representation properties underlying this experiment. The blind directions isolate the information absent from each modality, while the factor-effect measurements show that the missing quantities affect the target post-action response. The $AB/BA$ difference measures the effect of action order after the inputs stop.

The learned-model diagnostics use the 30,000-update convergence analysis. Oracle NMSE evaluates execution from exact response coordinates relative to the corresponding zero-coordinate prediction. The fused-information eigenvalue describes the directions constrained by the learned evidence factors; posterior trace measures the remaining coordinate uncertainty after combining them with the prior. Model performance comparisons appear in Figure~\ref{fig:ordered} and Table~\ref{tab:orq}.

\begin{table}[H]
\caption{\textbf{Physical and representation diagnostics of the controlled system.}
Blind-direction values are maximum evidence variations over the indicated parameter, with the other parameters fixed. Factor-effect RMS compares adjacent parameter levels in normalized test responses. Learned-model diagnostics use the 30,000-update training budget, with the executor selected by development error and fusion from the rank-two-factor model. The information eigenvalue excludes prior precision; trace reduction is relative to the smaller mean posterior trace of the two single modalities.}
\label{tab:controlled-diagnostics}
\centering
\small
\renewcommand{\arraystretch}{1.16}
\setlength{\tabcolsep}{5pt}
\begin{tabularx}{\linewidth}{@{}>{\raggedright\arraybackslash}p{0.29\linewidth}>{\raggedright\arraybackslash}p{0.22\linewidth}>{\raggedright\arraybackslash}X@{}}
\toprule
Diagnostic & Observed value & Interpretation\\
\midrule
Free-decay variation over $F_y$ & $0$ & Free decay is invariant to yield force.\\
\addlinespace[4pt]
Hysteresis variation over $c$ & $0$ & Settled hysteresis is invariant to damping.\\
\addlinespace[4pt]
Target-response RMS from changing $k,c,F_y$ & $k$: 0.09509\newline$c$: 0.03925\newline$F_y$: 0.09041 & All three physical parameters affect the test response.\\
\addlinespace[4pt]
Chart relative residual & $0.102265$ & Three coordinates retain $98.95\%$ of centered fitting-response variation.\\
\addlinespace[4pt]
Oracle NMSE, dev / test & $0.0756$ / $0.1844$ & Exact coordinates give lower error than the zero-coordinate prediction.\\
\addlinespace[4pt]
$AB/BA$ order-effect RMS & $0.62016$ & Changing action order changes the zero-input response.\\
\addlinespace[4pt]
Fused information $\lambda_{\min}$ & $0.5867$ & The evidence factors jointly constrain all three chart directions.\\
\addlinespace[4pt]
Mean posterior-trace reduction & $68\%$ & Fusion reduces coordinate uncertainty relative to either single modality.\\
\bottomrule
\end{tabularx}
\end{table}

\end{document}